\documentclass[sigconf]{acmart}

\AtBeginDocument{%
  \providecommand\BibTeX{{%
    \normalfont B\kern-0.5em{\scshape i\kern-0.25em b}\kern-0.8em\TeX}}}

\setcopyright{none}
\copyrightyear{none}
\acmYear{none}
\acmDOI{}
\acmConference[DSHealth '20]{DSHealth '20: 2020 KDD Workshop on Applied Data Science for Healthcare}{August 24, 2020}{San Diego, CA}
\acmBooktitle{}
\acmPrice{}
\acmISBN{none}

\usepackage[utf8]{inputenc} % allow utf-8 input
\usepackage[T1]{fontenc}    % use 8-bit T1 fonts
\usepackage{hyperref}       % hyperlinks
\usepackage{url}            % simple URL typesetting
\usepackage{booktabs}       % professional-quality tables
\usepackage{amsfonts}       % blackboard math symbols
\usepackage{nicefrac}       % compact symbols for 1/2, etc.
\usepackage{microtype}      % microtypography
\usepackage{appendix}
\usepackage{enumitem}
\usepackage{amsmath}

\usepackage{subcaption}
\usepackage{graphicx}
\usepackage{amsmath}
\usepackage{booktabs}
\usepackage{comment}
\usepackage{wrapfig}
\usepackage{color}   \usepackage{amssymb}

\newcommand{\model}{\textsf{EBmRNN}}
\newcommand{\compmodel}{\textsf{EmRNN}}

\newcommand{\relu}{\text{ReLU}}
\newcommand{\softmax}{\text{Softmax}}
\newcommand{\gsoftmax}{\text{Gumbel-Softmax}}

\makeatletter
\def\@copyrightspace{\relax}
\makeatother
\begin{document}

%%
%% The "title" command has an optional parameter,
%% allowing the author to define a "short title" to be used in page headers.
\title{Explicit-Blurred Memory Network for Analyzing Patient Electronic Health Records}

% The \author macro works with any number of authors. There are two commands
% used to separate the names and addresses of multiple authors: \And and \AND.
%
% Using \And between authors leaves it to LaTeX to determine where to break the
% lines. Using \AND forces a line break at that point. So, if LaTeX puts 3 of 4
% authors names on the first line, and the last on the second line, try using
% \AND instead of \And before the third author name.
\author{Prithwish Chakraborty}
%\authornote{corresponding author}
\email{prithwish.chakraborty@ibm.com}
\affiliation{%
  \institution{Center for Computational Health, IBM Research}
  % \streetaddress{P.O. Box 1212}
  \city{Yorktown Heights}
  \state{NY, USA}
  \postcode{10598}
}

\author{Fei Wang}
\email{feiwang.cornell@gmail.com}
\affiliation{%
  \institution{Department of Healthcare Policy and Research, \\ Weill Cornell Medicine, Cornell University}
  % \streetaddress{P.O. Box 1212}
  \city{New York}
  \state{NY, USA}
  \postcode{10065}
}

\author{Jianying Hu}
\email{jyhu@us.ibm.com}
\affiliation{%
  \institution{Center for Computational Health, IBM Research}
  % \streetaddress{P.O. Box 1212}
  \city{Yorktown Heights}
  \state{NY, USA}
  \postcode{10598}
}

\author{Daby Sow}
\email{sowdaby@us.ibm.com}
\affiliation{%
  \institution{Center for Computational Health, IBM Research}
  % \streetaddress{P.O. Box 1212}
  \city{Yorktown Heights}
  \state{NY, USA}
  \postcode{10598}
}

%%
%% By default, the full list of authors will be used in the page
%% headers. Often, this list is too long, and will overlap
%% other information printed in the page headers. This command allows
%% the author to define a more concise list
%% of authors' names for this purpose.
% \renewcommand{\shortauthors}{Trovato and Tobin, et al.}
\begin{abstract}
In recent years, we have witnessed an increased interest in temporal modeling
of patient records from large scale Electronic Health Records (EHR). While
simpler RNN models have been used for such problems, memory networks, which
in other domains were found to generalize well, are underutilized.
Traditional memory networks involve diffused and non-linear operations where
influence of past events on outputs are not readily quantifiable. We posit
that this lack of interpretability makes such networks not applicable for EHR
analysis. While networks with explicit memory have been proposed recently,
the discontinuities imposed by the discrete operations make such networks
harder to train and require more supervision.  The problem is further
exacerbated in the limited data setting of EHR studies. In this paper, we
propose a novel memory architecture that is more interpretable than
traditional memory networks while being easier to train than explicit memory
banks. Inspired by well-known models of human cognition, we propose
partitioning the external memory space into (a) a primary explicit memory
block to store exact replicas of recent events to support interpretations,
followed by (b) a secondary blurred memory block that accumulates salient
aspects of past events dropped from the explicit block as higher level
abstractions and allow training with less supervision by stabilize the
gradients. We apply the model for 3 learning problems on ICU records from the
MIMIC III database spanning millions of data points. Our model performs
comparably to the state-of the art while also, crucially, enabling ready
interpretation of the results.
\end{abstract}

%%
%% The code below is generated by the tool at http://dl.acm.org/ccs.cfm.
%% Please copy and paste the code instead of the example below.
%%


%%
%% Keywords. The author(s) should pick words that accurately describe
%% the work being presented. Separate the keywords with commas.

%% A "teaser" image appears between the author and affiliation
%% information and the body of the document, and typically spans the
%% page.
%% \begin{teaserfigure}
%%   \includegraphics[width=\textwidth]{sampleteaser}
%%   \caption{Seattle Mariners at Spring Training, 2010.}
%%   \Description{Enjoying the baseball game from the third-base
%%   seats. Ichiro Suzuki preparing to bat.}
%%   \label{fig:teaser}
%% \end{teaserfigure}

%%
%% This command processes the author and affiliation and title
%% information and builds the first part of the formatted document.
\maketitle

\section{Introduction}\label{sec:introduction}

In this new era of Big Data, large volumes of patient medical data are
continuously being collected and are becoming increasingly available for
research.  Intelligent analysis of such large scale medical data can 
uncover valuable insights complementary to existing medical
knowledge and improve the quality of care delivery.
Among the various kinds of medical data available, longitudinal Electronic
Health Records (EHR), that comprehensively capture the patient health
information over time, have been proven to be one of the most important
data sources for such studies. EHRs are routinely collected
from clinical practice and the richness of the information they contain
provides significant opportunities to apply AI techniques to extract nuggets of
insight. Over the years, many researchers have postulated various temporal
models of EHR for tasks such as early identification of heart failure~\cite{sun2012combining},
readmission prediction~\cite{xiao2018readmission}, and
acute kidney injury prediction~\cite{tomavsev2019clinically}. 
For such analysis to be of practical use, the models should provide support for
generating interpretations or post-hoc explanations. 
While the necessary properties of interpretations / explanations are 
still being debated~\cite{jain2019attention}, it is generally desirable to 
ascertain the importance of past events on model predictions at a particular time point.
Furthermore, despite their initial success, RNN model applications for EHR also
suffer from the inherent difficult to identify and control the temporal contents
that should be memorized by these RNN models.

Contemporaneously, we have also witnessed tremendous architectural advances for temporal
models that are aimed at better generalization capabilities. In particular, memory networks
~\cite{NTMGraves,DNC,Sukhbaatar2015a} are an exciting class of architecture that
aim to separate the process of learning the operations and the operands 
by using an external block of memory to memorize past events from the data. 
Such networks have been extensively applied to different problems and were found to generalize well~\cite{DNC}. 
However, there have been only a limited
number of applications of memory networks for clinical data modeling~\cite{prakash2017condensed,le2018dual}.
One of the primary obstacle is the inherently difficult problem of identifying important past events due to the diffused manner in which such networks store past events in memories. 
While~\cite{TARDIS,TARDIS2} have explored the possibilities
of using explicit memories that can store past events exactly and have found 
varying degrees of success, such models are difficult to train. The discontinuities
arising from the discrete operations either necessitate learning with high levels of
supervision such as REINFORCE with appropriate reward shaping or are learned using 
stochastic reparameterization under annealing routines and deal with high variance in gradients.

In this paper, we propose {\model}: a novel explicit-blurred memory architecture
for longitudinal EHR analysis. Our model is inspired by the well-known 
Atkinson-Shiffrin model of human memory~\cite{Atkinson:1968}. 
Our key contributions are as follows:

\begin{itemize}[leftmargin=0.2in,noitemsep,nolistsep]
    
\item We propose a partitioning of external memory of generic memory networks
  into a blurred-explicit memory architecture that supports better interpretability
  and can be trained with limited supervision.

\item We evaluate the model over $3$ classification problems on longitudinal EHR data. 
  Our results show {\model} achieves accuracies comparable to state-of-the-art 
  architectures.

\item We discuss the support for interpretations inherent in {\model} and analyze
  the same over the different tasks.
\end{itemize}
\section{Methods}
\label{sec:methods}

\textbf{Model:} Memory networks are a special class of Recurrent Neural Networks 
that employ external memory banks to store computed results. The separation between
operands and operators provided by such architectures have been shown to increase
network capacity and/or help generalize over datasets. However, the involved operations
are in general highly complex and renders such networks very difficult to
interpret. 
Our proposed architecture is shown in Figure~\ref{fig:architecture}.
The architecture is inspired by the Atkinson-Shiffrin model of cognition and is composed
of three parts:
% \begin{wrapfigure}{r}{.4\textwidth}
\begin{figure}[htpb]
    \centering
    \includegraphics[width=\linewidth]{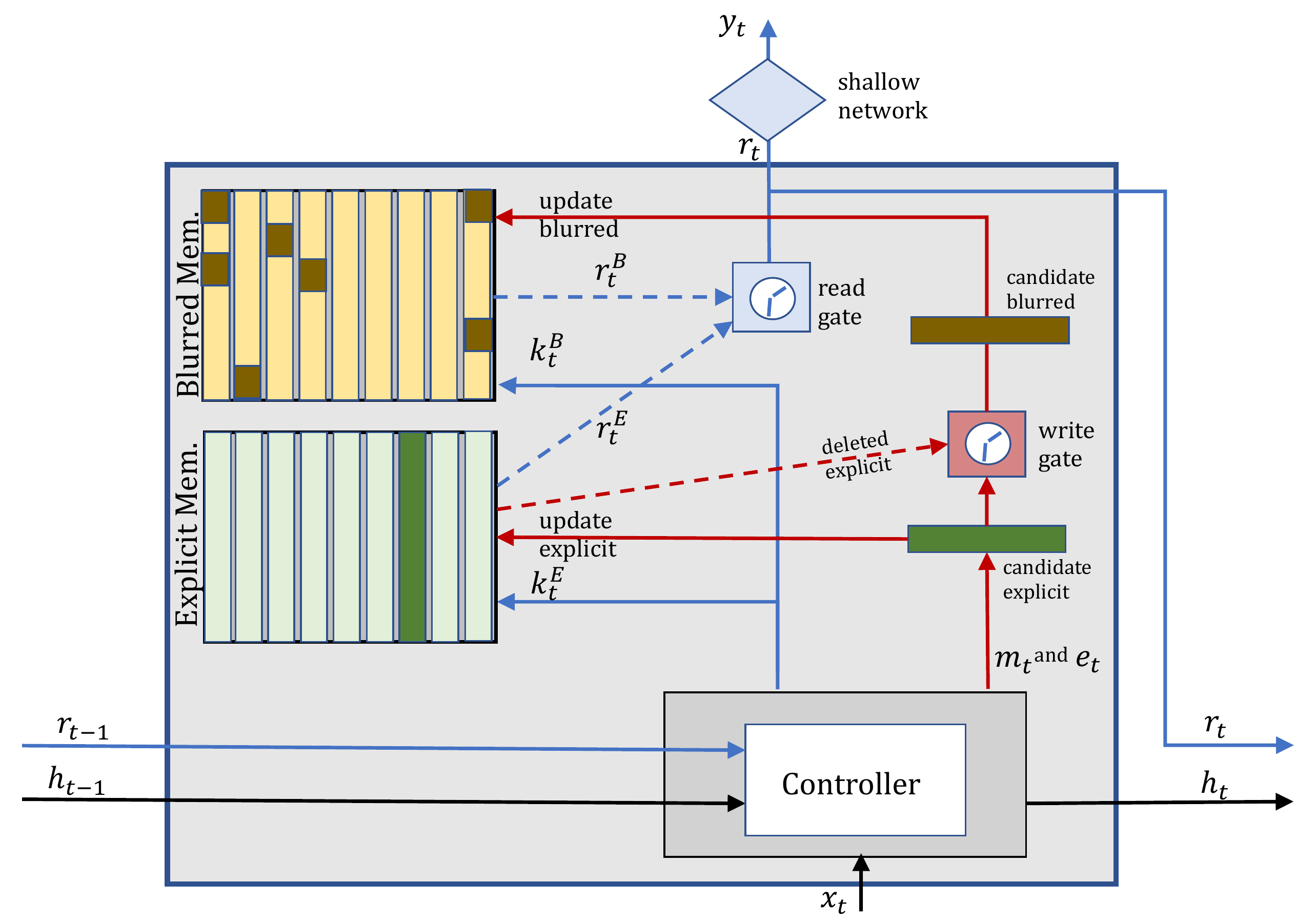}
    \caption{{\model} architecture: 
    Memory controller processes observations and can choose to store them
    discretely in explicit memory block or as  
    diffused higher level abstractions in blurred
    memory.
    }
\label{fig:architecture}
\end{figure}
% \end{wrapfigure}
% 
%
\begin{itemize}[leftmargin=0.2in,noitemsep,nolistsep]
\item a controller (e.g. a LSTM network) that processes inputs sequentially
  and produces candidate memory representation at each time point $t$ along
  with control vectors to manage the external memory. Mathematically, it can be expressed as follows:
\begin{equation}
  \label{eq:controller_udpate}
  \left[k_t^{E},k_t^{B},m_t,e_t\right],h_t  = RNN(h_{t-1}, x_t, r_{t-1})
\end{equation}

\item an `explicit' memory bank, where the generated candidate memory
  representation is stored. Depending on the outputs of a controlling read
    gate, the candidate memory can be stored explicitly or passed on to the
    blurred memory. When it is stored explicitly and the bank was already
    full, an older memory is removed based on the information content and
    passed on to the blurred memory bank. To update the memory explicitly, we 
    discretely select the index by make use of the Gumbel-Softmax trick as shown below:
\begin{equation}
  \label{eq:write_keys_usage}
  \begin{array}{lcl}
    u_t & = & \alpha_{\mathcal{E}} u_{t-1} + (1 - \alpha_{\mathcal{E}}) w_t^{r, \mathcal{E}} \\
    \gamma_t & = & \sigma\left(a_\gamma^T h_t + b_\gamma \right) \\
    w_t^{w, \mathcal{E}} & = & \gsoftmax\left( 1 - (\tilde{w}_t^{w, \mathcal{E}} + \gamma_t u_t) \right) \\
  \end{array}
\end{equation}
where, $u_t\in \mathbb{R}^D$ is a network learnt usage estimated.
$\alpha_{\mathcal{E}}$ is a hyper-parameter capturing the effect of current reads on 
the slots and $w_t$ is a one-hot encoded weight vector over memory slots. 
\end{itemize}

\begin{itemize}[leftmargin=0.2in,noitemsep,nolistsep]
\item The memory passed on to the blurred memory bank is diffused according to
  the control vectors and stored as high level concepts.
\end{itemize}

To generate outputs at time $t$, the architecture makes use of a read gate 
to select the memories stored in explicit and blurred memory that are 
useful at that time point. 
\begin{equation}
  \label{eq:read_gates}
  \begin{array}{lcl}
    g_t & = & \sigma\left(a^{\mathcal{B}T}_{g}r^{\mathcal{B}}_t + a^{\mathcal{E}T}_{g}r^{\mathcal{E}}_t + b_g\right) \\
    r_t & = & \relu\left((1 - g_t) \times W^{\mathcal{B}}_{r}r_t^{\mathcal{B}} + g_t \times W^{\mathcal{E}}_{r}r_t^{\mathcal{E}} + b_r\right) 
   
  \end{array}
\end{equation}
where, $\mathcal{B}$ and $\mathcal{E}$ are the blurred and explicit memories. 
$g_t \in \mathbb{R}$ is the read gate output and $r_t \in \mathbb{R}^D$ is the final 
output. The full model description is presented in the Appendix.

\textbf{Experimental setup:} We evaluated the performance of the proposed {\model} 
on the publicly available MIMIC III (Medical Information Mart for Intensive Care) 
data set~\cite{mimic3}.
% MIMIC III is a large, single-center database with data on patients admitted
% to critical care units at a large tertiary hospital. 
The data includes vital
signs, medications, laboratory measurements, observations and clinical notes.
For this paper, we focused on the structured data fields and followed the MIMIC
III benchmark proposed in~\cite{mimic3bench} to construct cohorts for $3$
specific learning tasks of great interest to the critical care community namely,
`In-hospital mortality', `decompensation',  and 'phenotype' classification. 
To estimate the effectiveness of the {\model} scheme, we compared it with the
following baseline algorithms: Logistic Regression using the features used in
\cite{mimic3bench}, Long Short Term Memory Networks, and Gated Recurrent Unit
(GRU) Networks. We also looked at a variant of {\model} that doesn't have access 
to blurred memory, hereby referred to as {\compmodel}. 
Comparison with {\compmodel} allows the training to proceed via a direct path to explicit memories and hence estimate 
its effect more accurately. 
{\compmodel} is completely interpretable while {\model} is interpretable to the limit allowed by the complexities of the problem. 
Details on the exact cohort definitions and
constructions are provided in \cite{mimic3bench}. More details on the tasks are also
presented in the Appendix.

\begin{table*}[hbtp]
    \centering
    \caption{Performance comparison for $3$ classification tasks on test dataset.}
    \label{tab:combined}
    % \scriptsize
	\begin{tabular}{lrrrr}
	\toprule
	{}        & In-Hospital mortality  & Decompensation & \multicolumn{2}{c}{Phenotype}  \\
	\cmidrule{4-5} 
	model     & AUC-ROC                & AUC-ROC        & AUC-ROC(macro) & AUC-ROC(micro)  \\       
	\midrule                                                                               
	LR        & 0.8485                 & 0.8678         & 0.7385       & 0.7995   \\      
	LSTM      & 0.8542                 & 0.8927         & {\bf 0.7670} & 0.8181   \\      
	GRU       & 0.8575                 & 0.8902         & 0.7664       & 0.8181   \\      
	EmRNN     & 0.8507                 & 0.8861         & {\bf 0.7670} & 0.8181   \\      
	EBmRNN    & {\bf 0.8612}           & {\bf 0.8989}   & 0.7598       & {\bf 0.8191}   \\
	\bottomrule
	\end{tabular}
\end{table*}

\begin{figure*}
	\centering
	\begin{subfigure}{0.33\linewidth} \includegraphics[width=\textwidth,
		height=0.15\textheight]{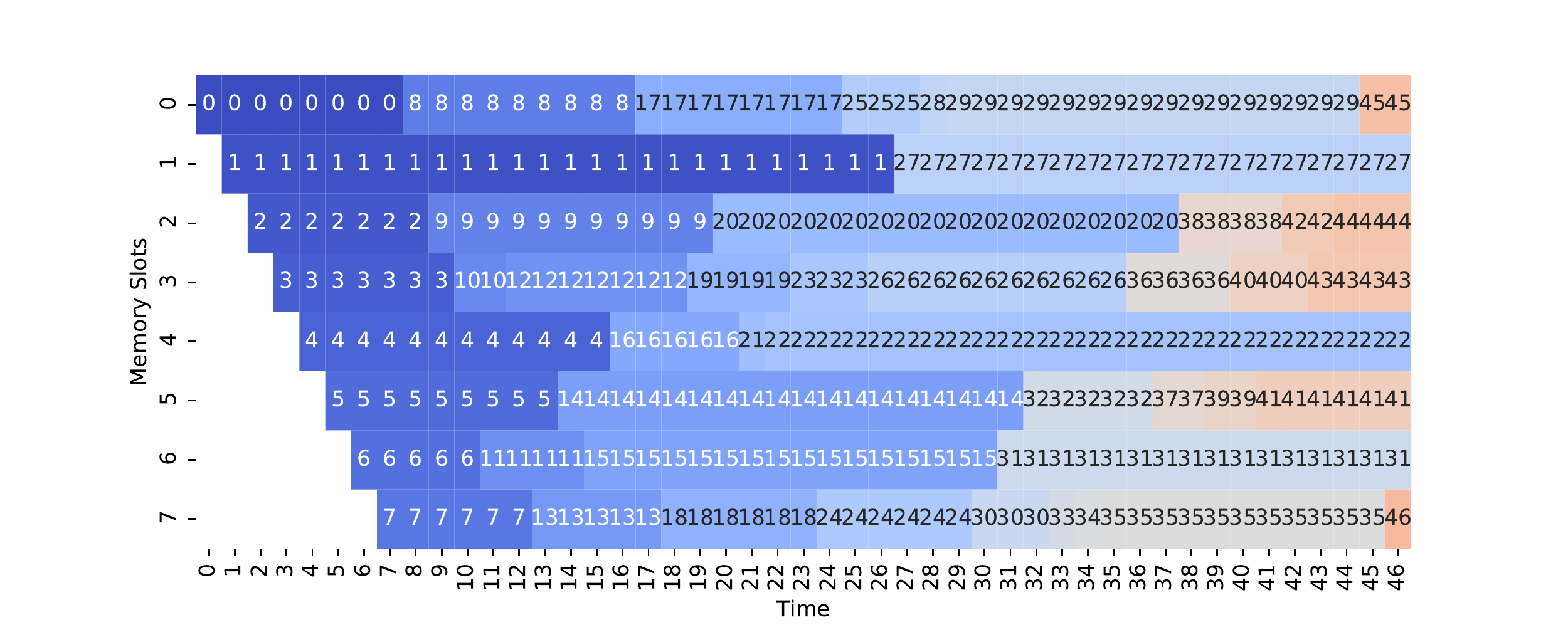}
		\caption{In Hospital Mortality} \end{subfigure}
\hspace{-1em}                                                                
	\begin{subfigure}{0.33\linewidth} \includegraphics[width=\textwidth, height=0.15\textheight]{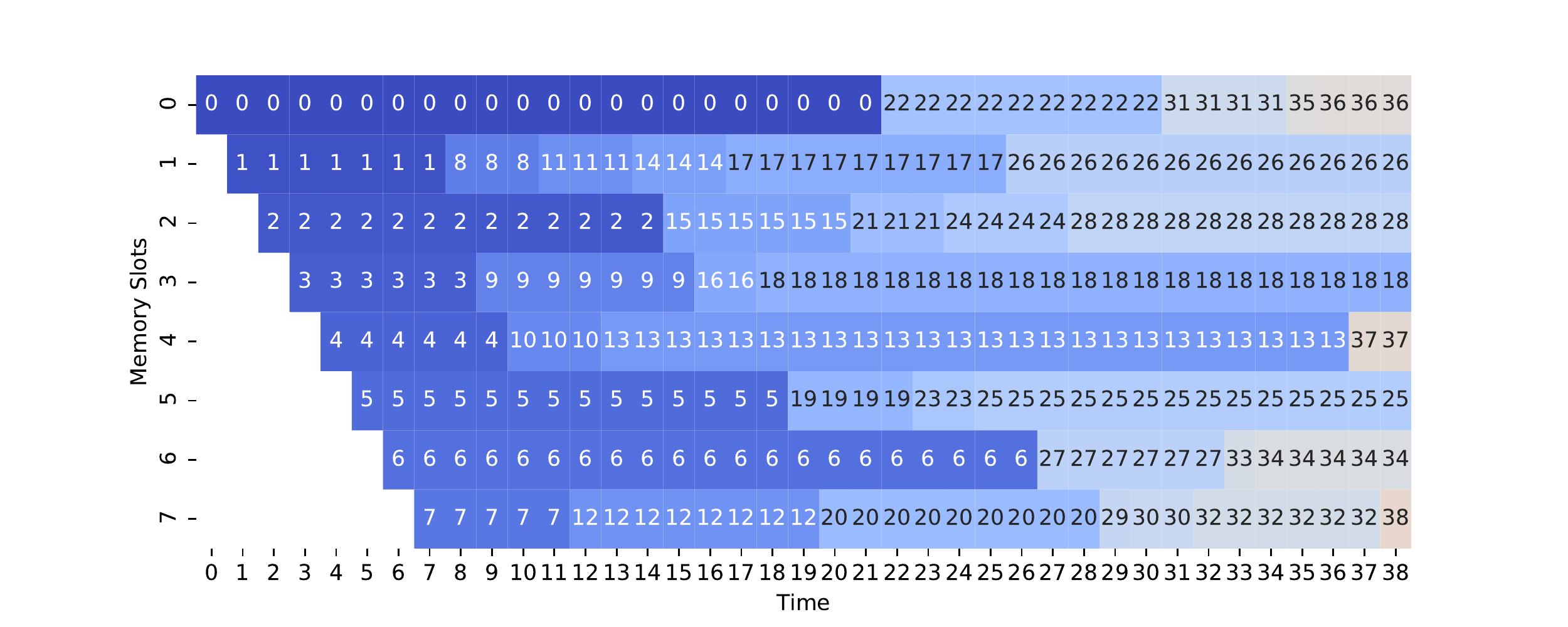}
		\caption{Decompensation} \end{subfigure}
\hspace{-1em}                                                                
	\begin{subfigure}{0.33\linewidth} \includegraphics[width=\textwidth, height=0.15\textheight]{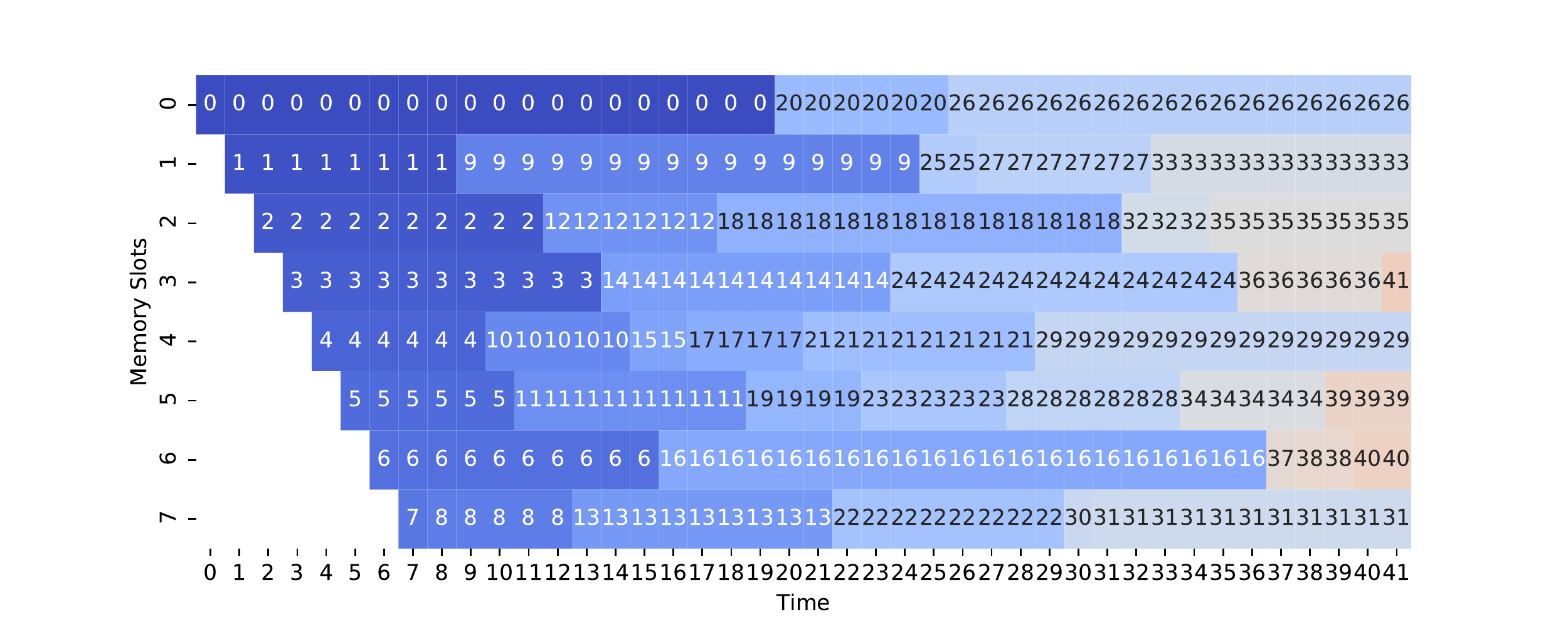}
		\caption{Phenotyping} \end{subfigure}
	\caption{Case Study: Explicit memory slot utilization to store events for $3$ separate patients for $3$ tasks using $8$ slots for the memory. Each slot is annotated by the time index of the event stored in memory. Memory utilization patterns exhibit long-term dependency modeling.} \label{fig:explicit_usage}
  % \vspace{-1em}
\end{figure*}
\begin{figure*}
	\centering
	\begin{subfigure}{0.33\linewidth} \includegraphics[width=\textwidth]{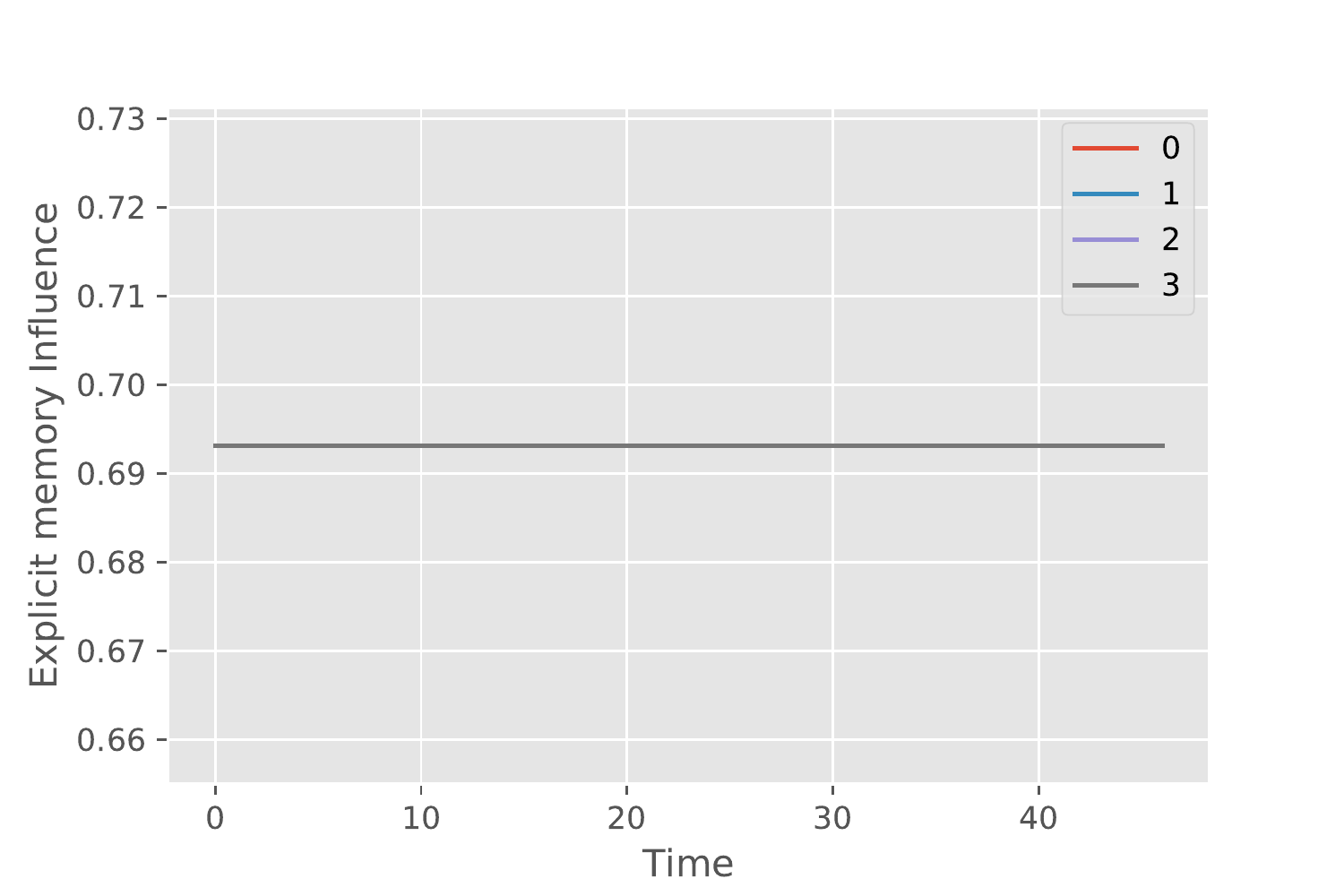}
		\caption{In Hospital Mortality} \end{subfigure}
\hspace{-1em}                                                                
	\begin{subfigure}{0.33\linewidth} \includegraphics[width=\textwidth]{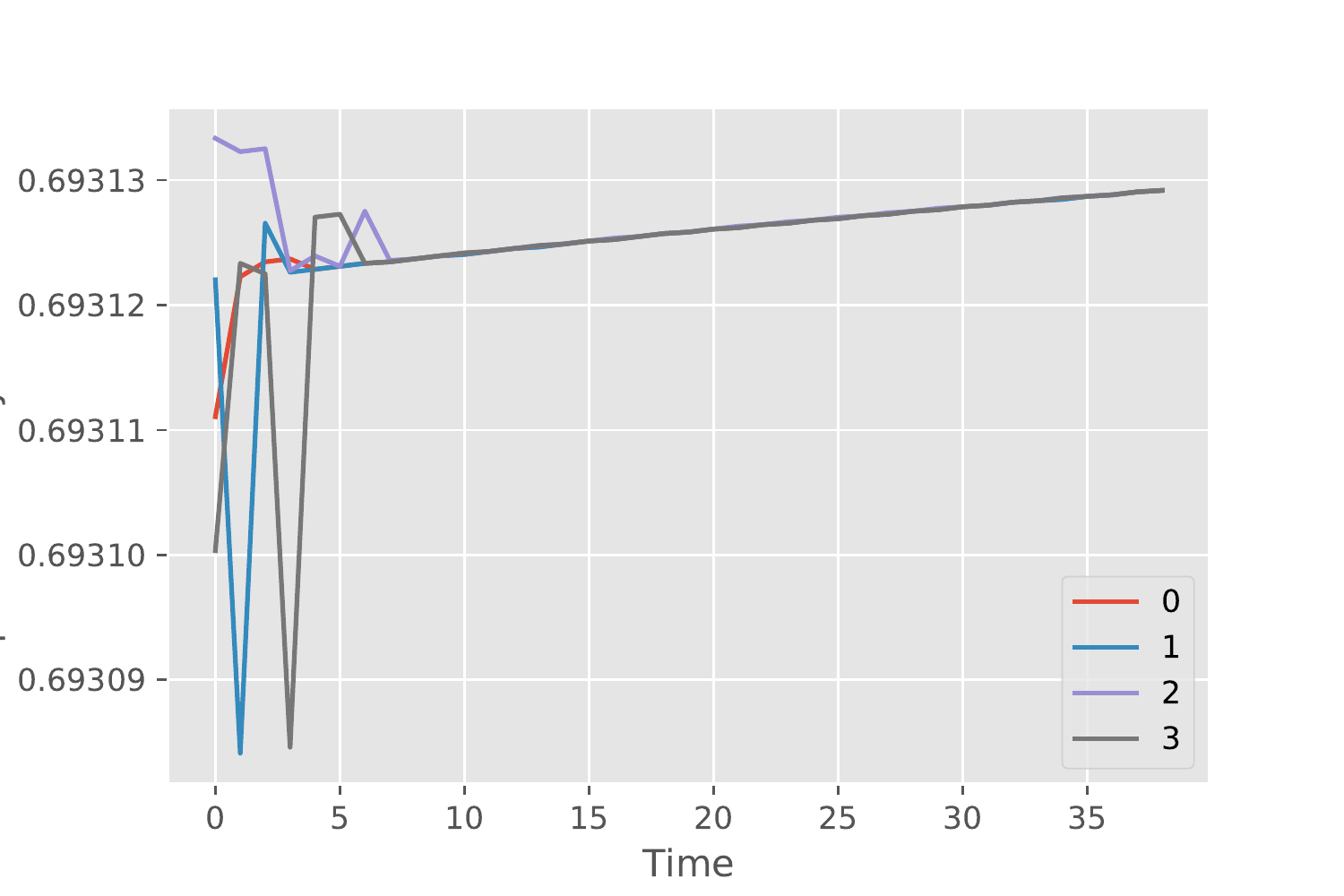}
		\caption{Decompensation} \end{subfigure}
\hspace{-1em}                                                                
	\begin{subfigure}{0.33\linewidth} \includegraphics[width=\textwidth]{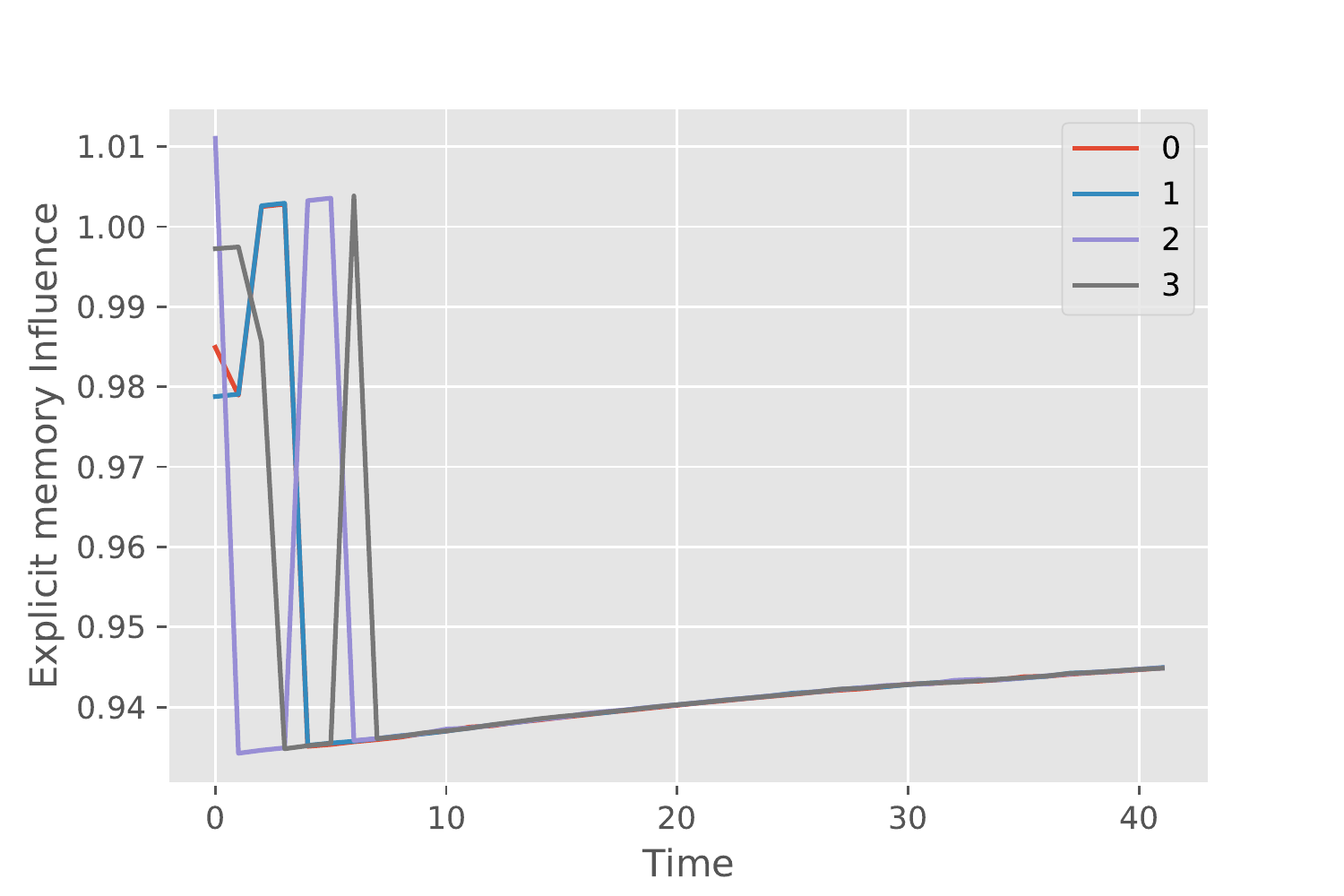}
		\caption{Phenotyping} \end{subfigure}
    \caption{Case Study: Influence of explicit memory for $3$ tasks, $1$
    patient per task, and $4$ hops of memory for read.  The legends indicate the explicit memory influence for each of the hops. Influence patterns vary
    across tasks indicating the task complexity as well as the modeling
    flexibility of {\model}.} \label{fig:explicit_influence}
  % \vspace{-1em}
\end{figure*}

\textbf{Data description:}
The dataset for each of the tasks is described below:
\\
\noindent
{\bf In Hospital Mortality Prediction:} This task is a classification
  problem where the learning algorithm is asked to predict mortality using the
    first 48 hours of data collected on the patient for each ICU stay. All ICU
    stays for which the length of stay is unknown or less than 48 hours have
    been discarded from the study. Following exactly the benchmark cohort
    constructions proposed in \cite{mimic3bench}, we were left with 17903 ICU
    stays for training and 3236 ICU stays for testing.\\
\\
\noindent
{\bf Decompensation Prediction:} This task is a binary classification
  problem. Decompensation is synonymous to a rapid deterioration of health
    typically linked to very serious complications and prompting ``track and
    trigger'' initiatives by the medical staff. There are many ways to define
    decompensation. We adopt the approach used in \cite{mimic3bench} to
    represent the decompensation outcome as a binary variable indicating
    whether the patient will die in the next 24 hours. Consequently, data for
    each patient is labeled every hour with this binary outcome variable. The
    resulting data set for this task consists of 2,908,414 training instances
    and 523,208 testing instances as reported in \cite{mimic3bench} with a
    decompensation rate of 2.06\\
\\
\noindent
{\bf Phenotyping:} This task is a multi label classification problem
  where the learning algorithm attempts to classify 25 common ICU conditions,
    including 12 critical ones such as respiratory failure and sepsis and 8
    chronic comorbidities such as diabetes and metabolic disorders. This
    classification is performed at the ICU stay level, resulting in 35,621
    instances in the training set and 6,281 instances in the testing set.

For each patient, the input data consists of an hourly vector of
features containing average vital signs (e.g., heart rate, diastolic blood
pressure), health assessment scores (e.g., Glasgow Come Scale) and various
patient related demographics.

\section{Results}
All the models were trained for 100 epochs. We used the recommended setting for the
baseline methods from~\cite{mimic3bench}. 
In this paper, we wanted to understand the relative importance of the memory 
banks and as such chose to study how the network uses the two different memory banks under similar capacity conditions.
For {\compmodel} and {\model}, the hyperparameters 
such as the memory size ($4-32$), controller hidden size ($4-32$), and the number of reads ($2-8$) were set using the validation set for each of the different datasets. 
While $\alpha_{\mathcal{E}}$ can be learned during the training process, 
following past work, we used a fixed value of $0.7$.
We chose a $2$-layered GRU with dropout as our controller and the models were trained using SGD with momentum ($0.9$) along with gradient clipping.
Table~\ref{tab:combined} shows the AUC-ROC for the different tasks.
Overall, we note that {\model} is on par or able to outperform
each of the baselines for each of the tasks.
Song {\em et al.}~\cite{song2018attend} found success with a
multi-layered large transformer based model and can be considered the state-of-the art including all architectures.
It is interesting to note that our results, using a single layer of memory, are comparable to the many-layered
transformer approach - thus indicating the efficiency of the proposed architecture. 
\noindent In the subsequent paragraphs, we discuss the key insights derived from 
the experiments.
\\
\\
\noindent
\textbf{How to interpret {\model}?} 
To analyze the interpretability inherent in
the model, we picked a patient for each of the tasks under consideration. 
We used a trained model with $8$ slots and allowing $4$ reads to generate the predictions.
As mentioned before, the explicit memory allows complete traceability of inputs
by storing each input in a distinct memory slot. 
Figure~\ref{fig:explicit_usage} depicts the contents of the explicit memory over 
time discretized by 1 hour.
Such slot utilization pattern provides an insight into the contents recognized 
by the network as being important for the task at hand. Furthermore, the plots
also exhibit that the model is able to remember, explicitly, far-off time points for an
extended period, before caching it into the blurred memory space.

\noindent \textbf{How to interpret the influence of explicit memory?} In addition
to exact memory contents, we can also analyze the importance of the explicit memory
for specific tasks by analyzing the control for the read gate $g_t$ over time.
Figure~\ref{fig:explicit_influence} shows the temporal progression of the read gate
for the $3$ patients from previous analysis for three distinct tasks. Interestingly,
we can see the model using different patterns of usage for different tasks. While
the network is assigning almost equal importance to both banks for in hospital mortality, 
it is placing high importance on explicit memory for phenotyping. This 
can also correlate with the improved performance for {\compmodel} for the phenotyping task.

\noindent \textbf{Why do we need the blurred memory?}  Given the interpretability provided
by the explicit memory, it may be tempting to avoid the use of blurred memory in favor 
of {\compmodel}. As our results indicate, such a model can perform well for certain
tasks. However, for tasks such as ``in-hosptial mortality'', the blurred memory 
provides the network with additional capacity.
Also, from a practical point 
of view, we found the {\compmodel} difficult to train where inspite of the 
Gumbel-Softmax reparameterization trick, the gradients frequently exploded 
and required higher supervision. On the other hand, the presence of the blurred
bank helped the training by providing a more tractable path. 
If the use case demands a higher value for intepretability,
we recommend to either use a smaller sized blurred memory bank or perform relative regularization
of the read gates for the blurred component. 

\section{Conclusion}
\label{sec:conclusion}
In this work, we have introduced {\model}, a memory network
architecture able to mimic the human memory models by combining
sensory, explicit and long term memories for classification tasks.
The proposed scheme achieves state-of-the-art levels of
performances while being more interpretable, especially when explicit memories 
are utilized more.
Our future work will aim at presenting such interpretations 
via end-to-end system following a user centered design approach.

% \clearpage
% \bibliographystyle{ACM-Reference-Format}
% \bibliography{references}
% \input{combined.bbl}
%%% -*-BibTeX-*-
%%% Do NOT edit. File created by BibTeX with style
%%% ACM-Reference-Format-Journals [18-Jan-2012].

% uncomment the next two lines to remove appendix
% \end{document}
% \begin{document}

\clearpage
\appendix
\pagenumbering{Roman}
\appendixpage

\section{Model Description}
\subsection{Explicit-Blurred Memory Augmented RNN}\label{sub:updated_model}
Let us denote the sequence of
observations as $x = x_1, x_2, \cdots, x_\mathcal{T}$, where $\mathcal{T}$ is the length of the
sequence and $x_t \in \mathbb{R}^{U}$. Similarly, let us denote the set of desired outputs as $y = y_1, y_2, \cdots, y_\mathcal{T}, y_t \in \mathbb{R}^{V}$. To model $y$ from $x$, $x$ is fed
sequentially to the proposed {\model} with parameters and hyper-parameters that will be defined below.

In {\model}, we split the conventional memory network architecture into two banks: (a) an explicit
memory bank ($\mathcal{E}$) and (b) a blurred or diffused memory bank($\mathcal{B}$). 
Figure~\ref{fig:architecture} shows a high level overview of the {\model} cell at time
$t$. This cell has access to an explicit memory bank $\mathcal{E}\in \mathbb{R}^{N^{\mathcal{E}}\times D}$ to persist past events
discretely. $N^{\mathcal{E}}$ denotes the capacity of the memory and $D$ is the dimensionality of each memory slot.  
This cell also has access to a blurred or diffused memory $\mathcal{B}\in \mathbb{R}^{N^{\mathcal{B}}\times D}$ where abstractions of important salient features from past observations are stored.

Observations at time $t$ are fed to this recurrent cell to produce an output $r_t \in \mathbb{R}^D$ based on the current input $x_t$, the external
explicit and blurred memories $\mathcal{E}$ and $\mathcal{B}$. $r_t$ summarizes information extracted from both $\mathcal{E}$ and $\mathcal{B}$ that is deemed relevant for the generation of the output $y_t$. $r_t$ is designed to contain enough abstraction of past observations seen by {\model}, including the current input $x_t$ so that specific tasks can generate a desired $y_t$ using only a shallow network outside of the cell. This design choice helps the interpretability of the model as it facilitates linking $y_t$ to memories in $\mathcal{E}$ pointing explicitly to inputs $x_t$, while still retaining the expressiveness of a blurred
memory. Analyzing how {\model} is using $\mathcal{E}$ provides a natural way to track how attentive {\model} is to input data stored in $\mathcal{E}$ while analyzing {\model}'s focus on $\mathcal{B}$ enables us to track the importance of long term dependencies. Details on how $r_t$ is computed are presented in the next subsection.

In addition to $\mathcal{E}$ and $\mathcal{B}$, there are three primary components controlling the functioning of the cell:

\begin{enumerate}
  \item The controller ($\mathcal{C}$), that senses inputs to {\model} and maps these inputs into control signals for the management of all read and write operations to the memory banks. 
  \item The read gate controlling read accesses to the memory banks from control signals emitted by the controller. 
  \item The write gate controlling writes into the memory banks from control signals emitted by the controller.
\end{enumerate}

In the remainder of this section, we describe these three components in details. 
\subsubsection{The Controller}\label{ssubcontrol}
At each time point $t$, the controller receives the current input $x_t$ and generates several outputs to manage $\mathcal{E}$ and $\mathcal{B}$ with appropriate read and write instructions sent to the read and write gates. 
As it receives $x_t$, the controller updates its hidden state $h_t\in\mathbb{R}^C$ based on
the past output of the cell $r_{t-1}$ , its past hidden state $h_{t-1}$ and current input $x_t$. In addition to updating its hidden state $h_t$, the controller emits two keys $k_t^{\mathcal{E}}\in\mathbb{R}^{D}$ and $k_t^{\mathcal{B}}\in\mathbb{R}^{D}$ to be used by the read gate to control access to memory contents from $\mathcal{E}$ and $\mathcal{B}$. To control write operations, the controller also produces $m_t \in \mathbb{R}^D$ a representation of the $x_t$ that will be consumed by the write gate. $m_t$ represents information from $x_t$ that is candidate for a write into $\mathcal{E}$ and $\mathcal{B}$. The controller also produces $e_t \in \mathbb{R}^D$, an erased weight vector that will be consumed by the write gate to forget content from $\mathcal{B}$. 
In this work, we model the controller with standard recurrent neural network architectures such as Gated Recurrent Units (GRU) or Long Short Term Memory networks (LSTM). The operations of the controller are summarized below: 
\begin{equation}
  \label{eq:controller_udpate}
  \left[k_t^{E},k_t^{B},m_t,e_t\right],h_t  = RNN(h_{t-1}, x_t, r_{t-1})
\end{equation}

\subsubsection{The Read Gate and Read Operations}
\label{ssub:read_op}
The read gate enforces read accesses from $\mathcal{E}$ and $\mathcal{B}$ by consuming $k_t^{\mathcal{E}}$ and $k_t^{\mathcal{B}}$ and comparing these keys against the 
content of the two memory banks $\mathcal{E}$ and $\mathcal{B}$. Using this addressing scheme, the following weight vectors over the memories are computed as follows:

\begin{equation}
  \label{eq:read_keys}
  \begin{array}{lcl}
    w_t^{r, {\mathcal{B}}} & = & \softmax(S(k^{\mathcal{B}}_t, \mathcal{B}_{t - 1}))   \\
    w_t^{r, {\mathcal{E}}} & = & \gsoftmax(S(k^{\mathcal{E}}_t, \mathcal{E}_{t -1}))   \\
  \end{array}
\end{equation}
where $S$ denotes an appropriate distance function between the key vectors and
the memory locations. For our purpose, we use the cosine similarity measure as a distance
function. $w_t^{r, {\mathcal{B}}} \in \mathbb{R}^{N^{\mathcal{B}}}$ and
$w_t^{r, {\mathcal{E}}} \in \mathbb{R}^{N^{\mathcal{E}}}$.
To ensure discrete access, $w_t^{r,{\mathcal{E}}}$ weights are required to be one-hot encoded vectors.
While Softmax is a natural choice for soft selection of indices for $w_t^{r, {\mathcal{B}}}$, its use is not 
applicable for the hard selection required for $w_t^{r,{\mathcal{E}}}$.
Gumbel Softmax is a newer paradigm that is applicable in this context compared to alternatives like top-K
Softmax that can introduce discontinuities. Gumbel Softmax uses a stochastic re-parameterization scheme to
avoid non-differentiablities that arise from making discrete choices during normal model training.
We use the straight-through optimization procedure that allows the network to make
discrete decisions on the forward pass while estimating the gradients on the backward pass using Gumbel Softmax. 
More details on this scheme can be found from~\cite{TARDIS}.

The read vectors $r^{\mathcal{E}}_t$ and $r^{\mathcal{B}}_t$ from each of the banks are computed as follows:
\begin{equation}
  \label{eq:read_vectors}
    r^{\mathcal{B}}_t  = w_t^{r, \mathcal{B}}\mathcal{B}_{t - 1}  \quad\quad
    r^{\mathcal{E}}_t  = w_t^{r, \mathcal{E}}\mathcal{E}_{t - 1} 
\end{equation}
$r^{\mathcal{B}}_t$ and $r^{\mathcal{E}}_t$ belong both to $\mathbb{R}^{D}$.
We combine the two content reads from the two banks using a gate as follows:

\begin{equation}
  \label{eq:read_gates}
  \begin{array}{lcl}
    g_t & = & \sigma\left(a^{\mathcal{B}T}_{g}r^{\mathcal{B}}_t + a^{\mathcal{E}T}_{g}r^{\mathcal{E}}_t + b_g\right) \\
    r_t & = & \relu\left((1 - g_t) \times W^{\mathcal{B}}_{r}r_t^{\mathcal{B}} + g_t \times W^{\mathcal{E}}_{r}r_t^{\mathcal{E}} + b_r\right) 
   
  \end{array}
\end{equation}
$g_t \in \mathbb{R}$ while $r_t \in \mathbb{R}^D$.
The final output from {\model} can then be
produced from a shallow layer that combines the contribution from the two
memory banks represented by $r_t$: 
\begin{equation}
\label{eq:output}
 y_t  =  \softmax\left(W_yr_t + b_y\right)
\end{equation}

Equation~\ref{eq:read_gates} ensures that the network can learn to produce its desired output $y_t$ using information from either memory banks. The gated value $g_t$ controls the relative effect of the blurred and
explicit memories on the output. On one hand, higher average values of $g_t$ would ensure that the network relies more on explicit memories and be as such easier to interpret. On the other hand, lower values of $g_t$ causes the network to rely more on blurred memories and be harder to interpret. Depending on the learning task at hand, there could be an interesting trade-off between learning performance and interpretability that can be controlled by this gating scheme. In fact, one could introduce a hyper-parameter in \ref{eq:read_gates} to control this trade-off between $W^{\mathcal{B}}_r$ and $W^{\mathcal{E}}_r$. 

The read operations are repeated $K$ times to generate $K$ hops from the memory.

\subsubsection{The Write Gate and Write Operations}\label{ssub:write_op}

Once memories are read, the controller updates the memory banks
for the next state. 
At each time point, the controller generates the memory representations, 
$m_t$, for the input $x_t$. 
The update strategy for the two banks are slightly different,
and we start by describing the explicit bank update first.

\par\noindent
\textbf{\textit{Explicit memory update:}}
As long as the explicit bank is not full, newer memories $m_t$ are simply appended to it and the update equation can be given as:
\begin{equation}
  \label{eq:write_basic}
  \mathcal{E}_t = [\mathcal{E}_{t-1}; m_t]
\end{equation}

Once the entire memory is filled up, the network needs to learn to forget
less important memory slots to generate a filtered explicit memory
$\tilde{\mathcal{E}_{t - 1}}$ and update the memory following equation~\ref{eq:write_basic}.
From an information theoretic intuition, more information can be retained by the network
by sustaining a higher entropy within the memory banks. 
The network learns the importance of the old memories with respect to new memory candidate content $m_t$ as follows:

\begin{equation}
  \label{eq:write_keys_content}
  \begin{array}{lcl}
    \tilde{w}_t^{w, {\mathcal{E}}} & = & \softmax(1 - S(m_t, \mathcal{E}_{t -1})) \\
  \end{array}
\end{equation}

$\tilde{w}_t^{w, {\mathcal{E}}}\in \mathbb{R}^{N^{\mathcal{E}}}$.
Equation~\ref{eq:write_keys_content} only uses the content to generate the
importance of the memory locations. Specifically, interpreting these values of
$\tilde{w}_t^{w, {\mathcal{E}}}$ in terms of retention probabilities, locations
with dissimilar contents will have higher retention probability - thereby
forcing the network to store discriminative content in the explicit memory.

Past research has also shown that usage-based addressing can significantly
improve the expressiveness of the network. We follow the scheme proposed
by~\cite{TARDIS} and make use of an auxiliary variable $u_t$ that tracks a
moving average of past read values for each memory locations of $\mathcal{E}$.
The final write vector along with all the usage update is given as:

\begin{equation}
  \label{eq:write_keys_usage}
  \begin{array}{lcl}
    u_t & = & \alpha_{\mathcal{E}} u_{t-1} + (1 - \alpha_{\mathcal{E}}) w_t^{r, \mathcal{E}} \\
    \gamma_t & = & \sigma\left(a_\gamma^T h_t + b_\gamma \right) \\
    w_t^{w, \mathcal{E}} & = & \gsoftmax\left( 1 - (\tilde{w}_t^{w, \mathcal{E}} + \gamma_t u_t) \right) \\
  \end{array}
\end{equation}
$u_t\in \mathbb{R}^D, \gamma_t \in \mathbb{R}, w_t^{w, \mathcal{E}}\in \mathbb{R}^{N^{\mathcal{E}}}$.
$\alpha_{\mathcal{E}}$ is a hyper-parameter capturing the effect of current reads on 
the slots.

Although, other addressing mechanisms have been proposed in literature, we chose this setting for model simplicity and also to better capture the desirable properties of EHR applications.

The explicit bank $\mathcal{E}$ is then updated by removing the slot with the
highest value of $w_t^{w, \mathcal{E}}$ ($\hat{m}_t$ from slot $j$) and
replacing its content with $m_t$. At that time, we also reset the usage value
for the slot (i.e. $u_t[j] = 0$).

Similar to the read operations,  $w_t^{w, E}$ is a one-hot encoded vector, the
equations for the popped memory, and subsequently update of the explicit memory
are given as below:

\begin{equation}
\begin{array}{lcl}
    \hat{m_t} & = & \mathcal{E}_{t-1}w_t^{w, \mathcal{E}}\\
 \mathcal{E}_t & = &  \mathcal{E}_{t-1} \circ (1_{N^{\mathcal{W}}\times D} - w_t^{w, \mathcal{E}} 1_{D}) + w_t^{w, \mathcal{E}}m^{\mathcal{E}}_t \\
 \end{array}
\end{equation}
where  $1_{N^{\mathcal{W}}\times D}$ represents a $N^{\mathcal{W}}\times D$
matrix of all 1 and $1_D$ represents the same for a $D$ dimensional vector.

\noindent
\textbf{\textit{Blurred memory update}}: The Blurred memories are used to represent past events
with more abstract concepts that can capture long term dependencies. The memory bank $\mathcal{B}$ provides a place for memories forgotten from the explicit bank to be stored in more abstract sense. $\mathcal{B}$ also allows {\model} to track and access a higher dimensional construct of current memory representation.

We generate a candidate blurred memory using the following equation:
\begin{equation}
  \begin{array}{lcl}
    f_t  & = & \sigma(W^i_f m_t + W^{\mathcal{E}}_f \hat{m}_t + b_f) \\
    m^{\mathcal{B}}_t & = & \relu((1 - f_t) W_i m_t + f_t W_{\mathcal{E}} \hat{m}_t + b_m ) \\
  \end{array}
\end{equation}

We generate write-vectors $w_t^{w, \mathcal{B}}$ using a formulation similar to equation~\ref{eq:write_keys_content} by replacing the $\gsoftmax$ with a $\softmax$. The final 
update equation for the blurred memory can then be given as follows:

\begin{equation}
  \begin{array}{lcl}
    \tilde{w}_t^{w, {\mathcal{B}}} & = & \softmax(S(m^{\mathcal{B}}_t, \mathcal{B}_{t-1})) \\
    \mathcal{B}_t & = &  \mathcal{B}_{t-1} \circ (1_{N\times W} - w_t^{w, \mathcal{B}} e_t) + w_t^{w, \mathcal{B}}m^{\mathcal{B}}_t
  \end{array}
\end{equation}

where $e_t$ is an erase weight generated by the controller.

\end{document}